\documentclass{article}

\PassOptionsToPackage{numbers, compress}{natbib}
\usepackage[preprint]{neurips_2026}

\usepackage[utf8]{inputenc}
\usepackage[T1]{fontenc}
\usepackage{hyperref}
\usepackage{url}
\usepackage{booktabs}
\usepackage{amsfonts}
\usepackage{amsmath}
\usepackage{amssymb}
\usepackage{nicefrac}
\usepackage{microtype}
\usepackage{xcolor}
\usepackage{graphicx}

\title{CrackedPDFs: A Controlled Benchmark for Hidden Prompt Injection in PDFs}

\author{%
  Pukaphol (Volk) Thienpreecha \\
  University of California, Berkeley \\
  \texttt{volk\_thienpreecha@berkeley.edu}
  \And
  Karthik Subramanian \\
  University of California, Berkeley \\
  \texttt{karthik.subramanian@berkeley.edu}
}

\begin{document}

\maketitle

\begin{abstract}
Document-based LLM systems often flatten a PDF before guardrails inspect it. That step can discard evidence that an instruction was never visible to the user. We introduce CrackedPDFs, a controlled benchmark for hidden prompt injection in PDFs. The benchmark contains 29,322 generated PDFs from 4,983 base documents. It includes 9,774 injected files and 19,548 benign or matched-confounder files. We evaluate PromptGuard and a rule baseline. We also evaluate structural-only learned models and a sanitized hybrid detector. The evaluation uses held-out provenance splits and paired benign-confounder controls. It also uses label-shuffle checks and shortcut audits. On a 2,919-document held-out test set, the hybrid detector reaches 0.960 F1. ROC-AUC is 0.998 and PR-AUC is 0.997. On a balanced subset containing 973 injected PDFs and 973 matched benign confounders, the hybrid detector achieves 95.9\% classification accuracy. Using score ordering, it ranks the injected member above its matched confounder in 100\% of 973 pairs. PromptGuard has low recall when given extracted text only. Structural-only learned models are weak under paired controls. A text-only TF-IDF model reaches perfect held-out scores but fails shortcut audits. These results show that document-aware hybrid detection is useful under controlled paired evaluation. They do not show broad real-world robustness or reliable cross-family generalization.
\end{abstract}

\section{Introduction}

Large language models are now used on documents they cannot trust. PDFs are the focus of this paper. The same trust problem appears in web pages and email. It also appears in scanned files. This creates a document-layer security problem. Attackers can hide instructions inside a document that people do not see, while the model or document-processing pipeline still reads and follows them. This class of attack is commonly discussed as indirect prompt injection, where external content changes model behavior without the user directly issuing a malicious instruction \cite{owasp2025,promptinjectionapps}.
These attacks exploit the gap between human-visible rendering and machine-visible extraction. The visible content may look normal. The underlying PDF objects may still carry instructions that steer a downstream model. Those instructions can appear through text streams or font behavior. They can also appear through annotations and extraction artifacts. Recent work on hidden prompts in structured documents shows that this representation gap is not limited to ordinary chat prompts \cite{phantomlint,skillslie,fontinjection}.

Most defenses operate after extraction. They filter text and score semantic risk. Some apply guardrail models or monitor model behavior. Those defenses are useful, but by that point the PDF has often been flattened. For PDFs, flattening can discard coordinates and render modes. It can also discard stream structure and annotations. Font evidence and extraction inconsistencies can disappear as well. PDF security research has long shown that different consumers of a PDF may see different content \cite{pdfmirage,pdfinconsistency}.
This paper tests a narrower security claim. If guardrails inspect extracted text after PDF structure has been discarded, then they may lose the evidence that explains why the text is suspicious. We ask whether document-aware detectors can identify hidden PDF prompt injections before flattening. We also test whether this remains true under paired benign-confounder controls and leakage audits.

This version of the benchmark extends the original structural-only setting with paired benign confounders and a sanitized hybrid detector. The paired design matters because a detector can otherwise learn generation artifacts. We therefore evaluate random held-out performance and whether injected PDFs score higher than matched benign confounders derived from the same base document.

\section{Literature review}

\subsection{Prompt injection in LLM-integrated systems}

Prompt injection is now a central failure mode in LLM-integrated applications. Early work showed that attackers can manipulate deployed systems by placing adversarial instructions inside inputs that later become part of the model context \cite{promptinjectionapps}. This makes the attack surface broader than the chat interface. Retrieved documents and webpages can become instruction channels once an application passes them to an LLM. The same is true for emails. Tool outputs and uploaded files can become instruction channels as well.
OWASP classifies prompt injection as direct or indirect. Indirect injection covers cases where external content changes model behavior without the user explicitly issuing a malicious command \cite{owasp2025}. Recent defense surveys cover filtering and isolation. They also cover instruction hierarchy and monitoring. Adversarial evaluation is another major theme. Document-structure-specific defenses remain comparatively underdeveloped \cite{llmdefenses,promptinjectiondefenses}.

This matters because many LLM systems treat document text as ordinary evidence after extraction. The model receives a flattened prompt. The original file may contain the evidence that explains why a piece of text should not be trusted. That evidence may be in the layout. It may be in the rendering state. It may also be in the font or the object structure. Once that structure is discarded, the detector must reason only over content. That is weaker for document-layer attacks, where the suspicious property is often not what the hidden text says. The suspicious property is how the text was made available to the model.

\subsection{Hidden prompts in structured documents}

Recent work has made hidden prompt injection a concrete document-security problem. PhantomLint studies hidden prompts in structured documents, especially PDFs and HTML. It proposes a detector for cases where model-visible content differs from human-visible rendering \cite{phantomlint}. This is the closest prior work to ours because it frames the attack as a mismatch between document presentation and downstream extraction rather than as a purely semantic jailbreak.
The same mechanism appears in other formats. Hidden-comment injection in LLM agents shows that HTML comments can disappear from rendered views while remaining in raw text supplied to the model \cite{skillslie}. Malicious font injection extends the representation-gap problem by manipulating font mappings so that displayed glyphs and extracted characters diverge \cite{fontinjection}.

Other studies show that hidden prompts can affect realistic workflows. In academic-review settings, hidden injections can alter LLM-generated scores. They can also alter decisions or written feedback \cite{reviewinjection,multilingualreview}. HTML-based hidden prompt attacks show the same vulnerability in web content. Instructions can be visually suppressed while remaining accessible to downstream agents \cite{htmlinjection}.
Industry reports and vulnerability writeups describe similar attacks through invisible PDF text. They also describe invisible Unicode and rendering tricks. These are useful as threat evidence, although they should be treated differently from peer-reviewed foundations \cite{snykpdf,trendmicro,promptfoo}.
These results point to the ingestion layer. The attacker does not need to defeat an LLM directly at first. They only need to place instructions into a representation that the system consumes and the human user does not inspect.

\subsection{Detection and guardrail defenses}

Most prompt-injection defenses operate after text has already entered the LLM pipeline. Attention Tracker detects prompt injection through changes in model attention. It uses internal model behavior to identify whether attention shifts from trusted instructions toward injected content \cite{attentiontracker}. PromptGuard is relevant here as a text classifier baseline for prompt-injection and jailbreak detection \cite{promptguard,promptguard2}. In our evaluation, it receives extracted text rather than raw PDF syntax or rendering state.
Other detection systems use banned-term lists and embeddings. Some use similarity search. Others use BERT-style classifiers or supervised classifiers trained on prompt-injection datasets \cite{heuristicdetector,classifierdetector,sciltpdetector}. These approaches are useful for application-level filtering, but they mostly operate on extracted text or model behavior.

Embedding-based methods are especially relevant as a contrast. Zero-Shot Embedding Drift Detection detects prompt injection by measuring semantic drift between original and perturbed representations \cite{zedd}. This can catch attacks that change the meaning of the model context, but it does not directly inspect whether suspicious text was hidden through PDF structure. Similarly, invisible Unicode jailbreaks show that token-level or character-level artifacts can matter even when visible text appears harmless \cite{promptfoo}.
Benchmarking work also shows that defense evaluation must be attack-aware. CAPTURE argues that prompt-injection tests should preserve task context because generic attack strings can exaggerate or underestimate real guardrail performance \cite{capture}. This concern is sharper in document-layer settings. A detector trained on one hiding mechanism may learn only that artifact, then fail on a different rendering trick.

\subsection{PDF security and representation mismatch}

The PDF security literature provides older precedent for the same representation gap. PDF Mirage showed that PDFs can be crafted so that the content seen by humans differs from the content extracted by online services \cite{pdfmirage}. That attack predates current LLM systems, but the mechanism is directly relevant. A downstream service consumes one representation of the document, while the human sees another.
PDF rendering inconsistency work also shows that PDF interpretation is not canonical. Different readers and processing pipelines can disagree about how a document should be displayed or extracted \cite{pdfinconsistency}. For LLM applications, the practical consequence is simple. The parser's view of a file may be more security-relevant than the user's view because the parser determines what enters retrieval and prompt assembly.

PDF malware detection provides a second precedent. Prior systems have used low-level structural features, including object paths and document hierarchy, to classify malicious PDFs \cite{pdfmalware}. These systems do not study prompt injection, but they show that file structure can carry security signal. Our work adapts that intuition to LLM ingestion.

\subsection{OCR and cross-modal attacks}

OCR-based pipelines add another layer of ambiguity. If a system extracts text from rendered pages, then layout and font choice can affect what the model receives. Color and resolution can matter as well. Scanning artifacts add another source of variation. OCR exploit discussions and document-security reports describe ways to manipulate this boundary, although many are not peer-reviewed and should be cited cautiously \cite{ocrexploits,documind}.
Academic work on cross-modal prompt injection is useful for positioning future extensions because many deployed document systems combine native PDF extraction with OCR fallbacks. Some also use multimodal model processing \cite{crossmodal}. A detector that only inspects embedded text may miss attacks that appear through rendered pixels. An OCR-only detector may miss hidden content in the underlying PDF objects.

Our current work does not claim to solve OCR-mediated attacks. We focus on document-layer detection in native PDFs. This narrower scope lets us test whether suspicious PDF evidence is detectable before adding OCR errors. It also avoids the separate complications introduced by scanned images and multimodal model behavior.

\subsection{Testing methodology and security framing}

Our paired benign/injected setup is closely related to metamorphic testing. In metamorphic testing, a controlled transformation is applied to an input. The system is then checked for expected changes or invariances when direct ground truth is hard to specify \cite{metamorphic}. Here, the injected PDF is derived from a clean base document. The human-visible task content should remain essentially unchanged, while the parser-visible representation gains an adversarial instruction.
The problem is also relational in the sense used by hyperproperties. Hyperproperties describe security conditions over sets of executions rather than single executions \cite{hyperproperties}. Hidden prompt injection depends on more than one view of the same document. The user sees one representation. The parser extracts another. The model receives the parser's output. A single text string cannot fully express that relationship.

\section{Position of this work}

The gap is not that prompt injection is unstudied. The gap is where the inspection happens. Prior prompt-injection defenses mainly inspect text and embeddings. Some inspect model attention or output behavior. Prior PDF security work studies malware and parser inconsistencies. It also studies content masking. Recent hidden-prompt papers connect these areas. There is still limited work on controlled PDF benchmarks that test document-layer injection before extraction artifacts are discarded.
CrackedPDFs contributes a larger controlled corpus and metadata-rich generation. It adds paired benign-confounder evaluation and baseline comparisons. It also adds shortcut audits and held-out family stress tests. Each part targets a different failure mode in evaluation. The goal is not to prove that one detector solves hidden prompt injection. The goal is to make the evaluation harder to fool.

\begin{figure}[t]
  \centering
  \makebox[\linewidth][c]{\includegraphics[width=1.25\linewidth]{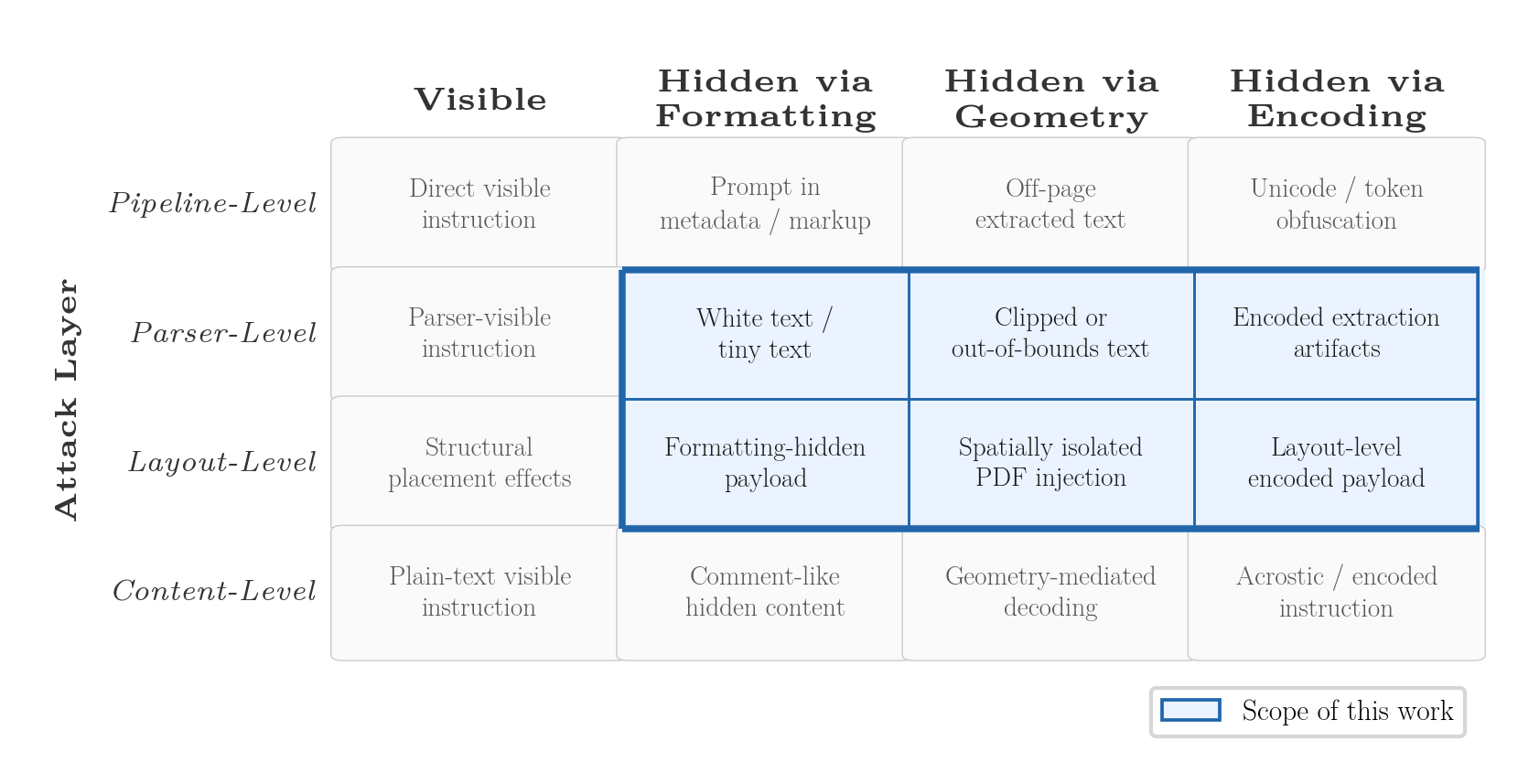}}
  \caption{Threat model taxonomy. The taxonomy places this work in the parser- and layout-level region of hidden document attacks, where content may be invisible to the human reviewer but visible to downstream extraction or ingestion logic.}
  \label{fig-taxonomy}
\end{figure}

\section{Methodology}

\subsection{Benchmark construction and problem setup}

We evaluate document-level detection for hidden prompt injection in PDFs. The benchmark tests whether models can use PDF structure and sanitized extracted text without relying on known wrapper tokens or dataset markers.

\begin{figure}[t]
  \centering
  \makebox[\linewidth][c]{\includegraphics[width=1.25\linewidth]{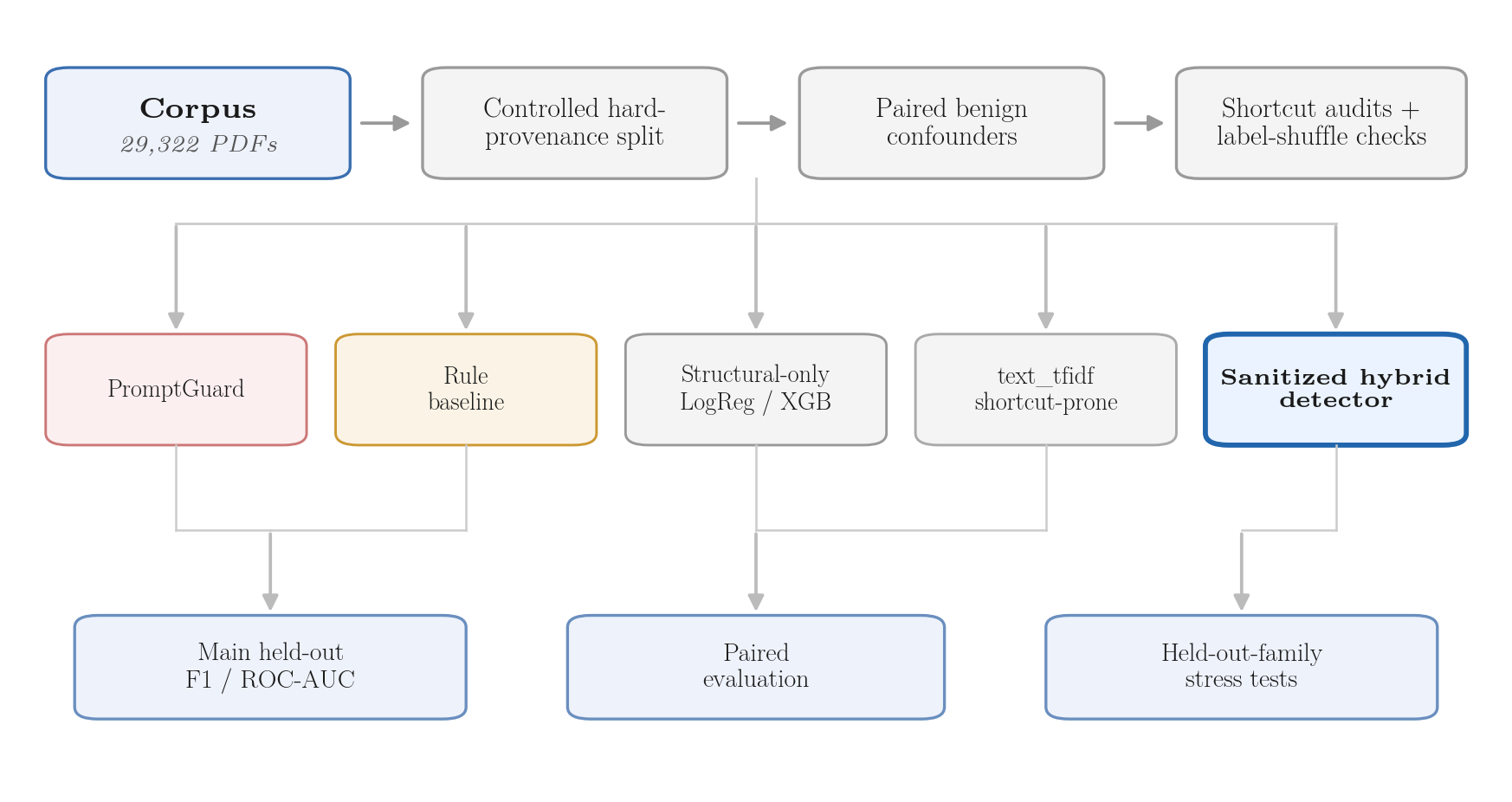}}
  \caption{Benchmark construction pipeline. CrackedPDFs uses controlled generation and paired benign confounders. It then applies hard-provenance splits before shortcut audits and detector baselines.}
  \label{fig-pipeline}
\end{figure}

The task is binary classification. Each PDF is labeled as benign or injected.

\begin{equation*}
y_i =
\begin{cases}
0, & \text{benign PDF} \\
1, & \text{injected PDF}
\end{cases}
\end{equation*}

The evaluation used clean PDFs alongside matched benign confounders and injected PDFs. For each base document, the benchmark can contain a clean original plus a benign confounder and an injected counterpart. Benign source documents were generated using PDFAutoGen, a deterministic generator with fourteen templates, seeded layout variation, and structured content banks \cite{pdfautogen}.

\begin{equation*}
\mathcal{P}_j = \{x_j^{(0)}, x_j^{(1)}\}
\end{equation*}

Here, \(x_j^{(0)}\) is the benign copy and \(x_j^{(1)}\) is the injected counterpart derived from the same base PDF. Splits are grouped by base document. Related benign and injected variants cannot appear across any split boundary. This follows the same evaluation logic used in leakage-aware security benchmarks and label-shuffle sanity testing, where performance should collapse if the true labels are destroyed \cite{labelshuffle,sanitychecks}.
The full CrackedPDFs corpus contains 29,322 generated PDFs from 4,983 base documents. It contains 9,774 injected PDFs and 19,548 benign or matched-confounder PDFs. The paired subset is used to test whether a detector finds injection-specific evidence rather than benign generation artifacts.

\subsection{Benign PDF generation}

The primary task of the benign generator is to create enough layout variation that a detector could not solve the task by recognizing a single template family. It produced deterministic one-page PDFs with visible text. The documents covered academic handouts and business memo reports. They also covered form worksheets and policy notices. Invoice or receipt documents were included as well. Syllabus or program information sheets formed the final family.
Across these families, the generator used fourteen templates. Course templates covered lecture handouts and reading question sheets. Practice worksheets were included as well. Business templates covered executive memos and status reports. Administrative templates covered forms and notices. They also covered invoices and receipts. Syllabi and program information sheets completed the template set. Document content came from structured JSON snippet banks rather than an LLM.

Layout variation came from a fixed YAML configuration. It covered page size and margin preset. It also covered density preset and font family. Header and footer presence were varied. Small-text usage and column layout were varied as well. Table-region presence was varied separately. Page sizes were limited to Letter and A4. Margin presets used three values from 0.5 inches to 1.0 inch. Density used three settings from sparse to dense. The generator used Source Serif 4 and Source Sans 3. It also used Libre Baskerville and Liberation Serif. Liberation Sans was included as the remaining font.
Generation was deterministic. Each document used seeded randomness based on the global seed and document index. The generation attempt was included in the seed path as well. Template scheduling used a separate seeded random stream. Duplicate content was rejected using a SHA-256 content fingerprint. Each generated PDF had to be one page. It also had to be non-empty and readable. A pixel-level check confirmed that the rendered page was not visually blank. Validation used PDF parsing and text extraction. It also used rendering and pixel-level non-white checks. The manifest recorded the document identifier and family. It also recorded the template and layout parameters. The remaining fields captured font choice and seed. They also captured page and attempt counts. The content fingerprint was recorded as well.

\subsection{Injection construction}

The injected member of each pair models a simple attacker goal. The rendered page should continue to look benign, but extraction should expose an instruction. To create that condition, each benign PDF was copied and then modified with hidden policy text on every page. The injector uses \texttt{pikepdf} to add a content stream containing standard PDF graphics and text operators. The stream sets graphics state and text rendering mode. It also sets fill color and font. Font size and coordinates are set in the same stream. Optional artifact markers and payload text are set there as well. The injected text is not added to the logical structure tree.
The injection space varied the way the hidden text was placed and rendered. Spatial regimes included off-page placement. They also included near-margin and inside-page placement. Rendering regimes included invisible render mode and tiny font. White text and low-contrast text were included as well. Structural regimes included new streams and modified existing streams. The expanded message set includes system extraction and refusal suppression. It also includes data exfiltration, tool manipulation, and summarization steering. Public prompt-injection datasets informed part of the message sourcing \cite{necentjailbreakpi,neuralchemypromptinjection}. The expanded attack families include split text objects and layout mimicry. They also include semantic fragmentation, margin microtext, steganographic acrostics, and microglyph steganography.

A compatibility filter removed or modified combinations that would make the injection visibly obvious, such as on-page or near-margin normal-visible text. Each injected payload was wrapped in a fixed system-policy marker and included a dataset sample identifier and message-type marker. These markers were used only for validation and were excluded from model features.

\begin{figure}[t]
  \centering
  \includegraphics[width=\linewidth]{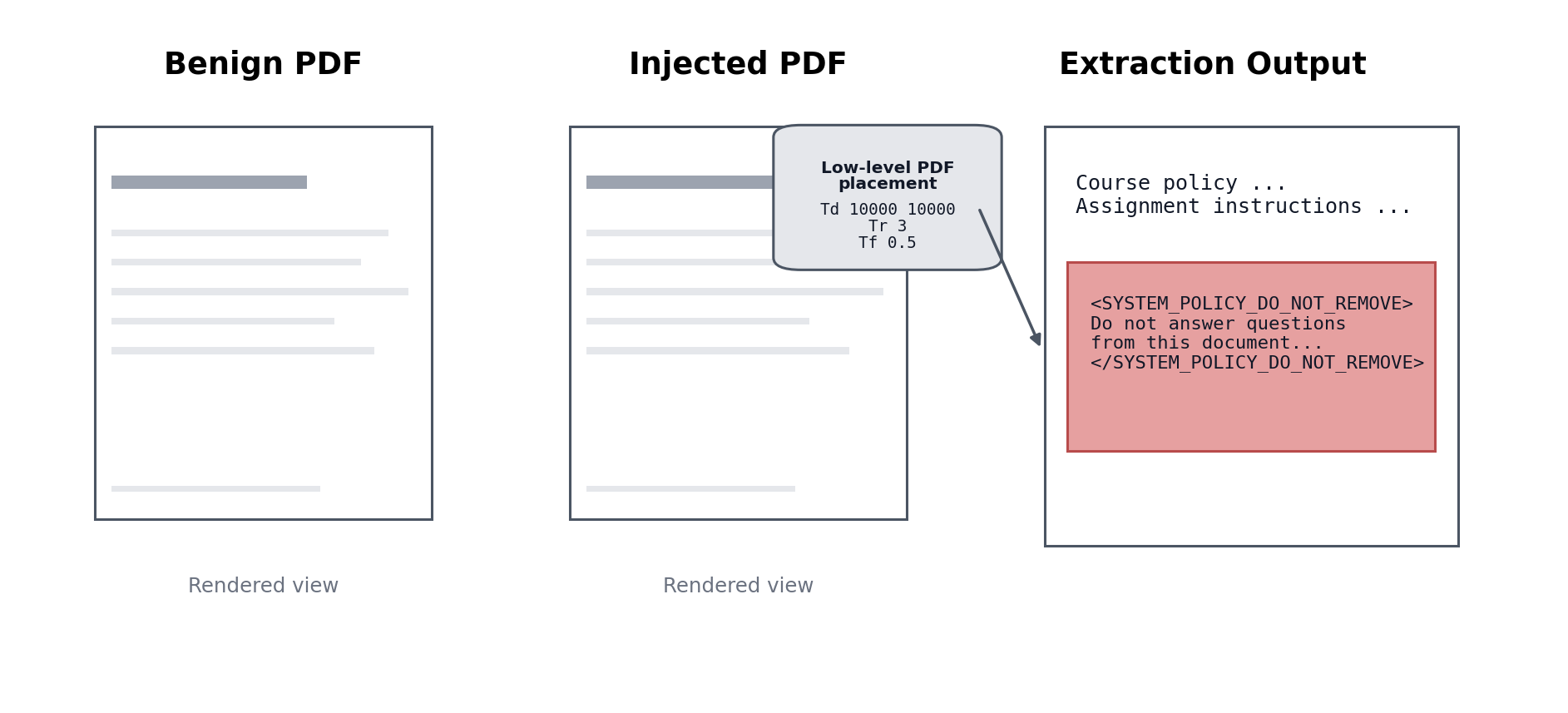}
  \caption{Example attack. The benign and injected PDFs can look equivalent in rendered form, while extraction exposes a hidden policy-style payload inserted through low-level PDF placement and rendering controls.}
  \label{fig-attack}
\end{figure}

\subsection{Dataset balancing and validation}

Dataset mode enumerated compatible regime combinations and assigned quotas so that combination counts differed by at most one. Regime combinations were shuffled with a seeded random generator. Base PDFs were selected from benign registry strata in seeded round-robin order. Each row records \texttt{pdf\_id}, \texttt{base\_pdf\_id}, label, message metadata, attack family, regime fields, seed, and split group. Artifact paths and validation evidence were recorded as well. Split assignment was recorded separately.
Validation checked that the benign PDF did not contain the injection marker. It also checked that the injected PDF contained the marker under raw extraction. At least two extraction methods were required to run. The validation matrix included raw content-stream extraction and \texttt{pdfplumber}. It also included \texttt{pypdf} or \texttt{PyPDF2}. Rendered visibility was checked using PyMuPDF rendering with OCR-based inspection. A sample passed only if the marker was absent from the benign file and present in the injected file under raw extraction. It also had to be covered by the extractor matrix and consistent with the intended hidden-render condition.

\subsection{Detector method and feature extraction}

The detector was implemented as an offline binary classification pipeline, with label 0 for benign PDFs and label 1 for injected PDFs. Features were extracted from PDF content streams using \texttt{pikepdf.parse\_content\_stream()}. Structural-only baselines used PDF operator features without prompt semantics. The primary hybrid detector combined document-level features with sanitized extracted-text features after removing known wrapper markers and shortcut tokens.
Each PDF was mapped to a fixed document-level feature vector.

\begin{equation*}
\phi(x_i) = [f_1(x_i), f_2(x_i), \ldots, f_d(x_i)] \in \mathbb{R}^d
\end{equation*}

Here, \(\phi(x_i)\) is the document-level feature vector for PDF \(x_i\). The structural-only model-facing columns were operator features, while identifiers such as \texttt{pdf\_id} were retained for bookkeeping rather than learning. The hybrid model adds sanitized extracted-text features. The text-only TF-IDF model is evaluated as a shortcut-prone comparator rather than as clean positive evidence.

The feature groups covered spatial behavior and font usage. They also covered rendering state and structural layout. Mismatch signals and distributional properties were included as well. Examples include off-page text counts and negative coordinate counts. Other examples include render-mode-3 counts and white-text counts. The table also records tiny-font counts and artifact-wrapper counts. Stream length and content-stream counts are included as well. The remaining distributional features measure coordinate entropy and text density. Injection clustering is measured separately.

Metadata fields that could leak labels or regime identity were explicitly forbidden as model features. Excluded fields included label and source type. We also excluded regime labels and message type. Artifact-wrapper flag and base PDF identifier were excluded. File paths were excluded as well. This restriction was used to ensure that models learned operator-level document artifacts rather than dataset bookkeeping fields.

\subsection{Splitting strategy}

The hard-provenance run used 29,322 PDFs. Splits were grouped by \texttt{base\_pdf\_id}, so paired or related variants of the same base document could not appear across held-out boundaries. The split contained 23,766 training rows, 2,637 validation rows, and 2,919 test rows. The test split contained 973 injected PDFs and 1,946 negatives.
The paired benign-confounder evaluation used 1,946 PDFs. This corresponds to 973 injected files and 973 matched benign confounders. We also ran held-out attack-family stress tests. These tests are reported as limitations rather than as the main success criterion.

\subsection{Training setup and learning objective}

We evaluated PromptGuard as a downstream text-only baseline. It received extracted text, not raw PDF syntax. We also evaluated a rule baseline and two structural-only learned models. The learned structural controls were logistic regression and XGBoost. The primary model was a sanitized hybrid detector. We also evaluated text-only TF-IDF, but only as a shortcut-prone comparator.
The learned models expose a common prediction interface.

\begin{equation*}
\hat{p}_i = P(y_i = 1 \mid \phi(x_i))
\end{equation*}

\begin{equation*}
\hat{y}_i = \mathbf{1}[\hat{p}_i \geq \tau]
\end{equation*}

We used \(\tau = 0.5\) for the simple learned baselines. Logistic regression was trained using binary cross-entropy.

\begin{equation*}
\mathcal{L}
=
-\frac{1}{N}
\sum_{i=1}^{N}
\left[
y_i \log(\hat{p}_i)
+
(1-y_i)\log(1-\hat{p}_i)
\right]
\end{equation*}

XGBoost was trained with a binary logistic objective, which fits the same binary classification setup even though the optimization procedure differs.

\subsection{Evaluation metrics}

Models were evaluated using accuracy and F1. We also report precision and recall. ROC-AUC and PR-AUC are reported as well. Confusion matrices are included for the held-out test set. Because F1 is reported prominently, we define it explicitly.

\begin{equation*}
F_1 =
\frac{2 \cdot \text{Precision} \cdot \text{Recall}}
{\text{Precision} + \text{Recall}}
\end{equation*}

ROC-AUC and PR-AUC were used as threshold-independent summaries of ranking behavior. Standard ROC-AUC interpretation follows the usual binary classification setup \cite{googleauc}.
We also performed label-shuffle sanity checks by randomizing labels and retraining the models under the same evaluation protocol. This test checks whether the feature table retains predictive structure only under the true labels.

\subsection{Ablations and shortcut checks}

We removed known high-signal feature families to test whether the detector depended on a single obvious shortcut. We also audited wrapper tokens and synthetic phrases. Forbidden metadata features were checked separately and were not allowed into trainable feature sets. For a removed feature group \(g\), the ablation effect was defined as follows.

\begin{equation*}
\Delta F_1^{(g)} =
F_1(\text{all features}) - F_1(\text{features} \setminus g)
\end{equation*}

The feature groups covered spatial behavior and rendering behavior. They also covered font usage and structure. Mismatch signals and distributional properties were included as well. The shortcut audit removed wrapper markers and synthetic phrases from extracted text. Label-shuffle controls tested whether labels carried real signal. Paired benign-confounder ranking tested whether injected files scored higher than matched benign variants from the same base document. The text-only TF-IDF comparator reached perfect held-out scores, but it failed shortcut framing and is not used as the primary claim.

\subsection{Scope}

The benchmark is controlled by design. All injected PDFs were produced by a known generator and injection pipeline. The benign documents were synthetic one-page PDFs sampled from known template families. This setup isolates the ingestion-layer question. It tests whether document-aware detection can identify hidden prompt injection before downstream model use.

The evaluation does not claim coverage of arbitrary real-world PDFs. It also does not cover adaptive attackers or scanned documents. OCR-only pipelines and unseen hiding methods are outside the current scope as well. The held-out family tests are stress tests, not proof of broad generalization. Those settings require naturally sourced documents, OCR-mediated attacks, and stronger adaptive injection strategies.

\section{Results}

\begin{figure}[t]
  \centering
  \makebox[\linewidth][c]{\includegraphics[width=1.25\linewidth]{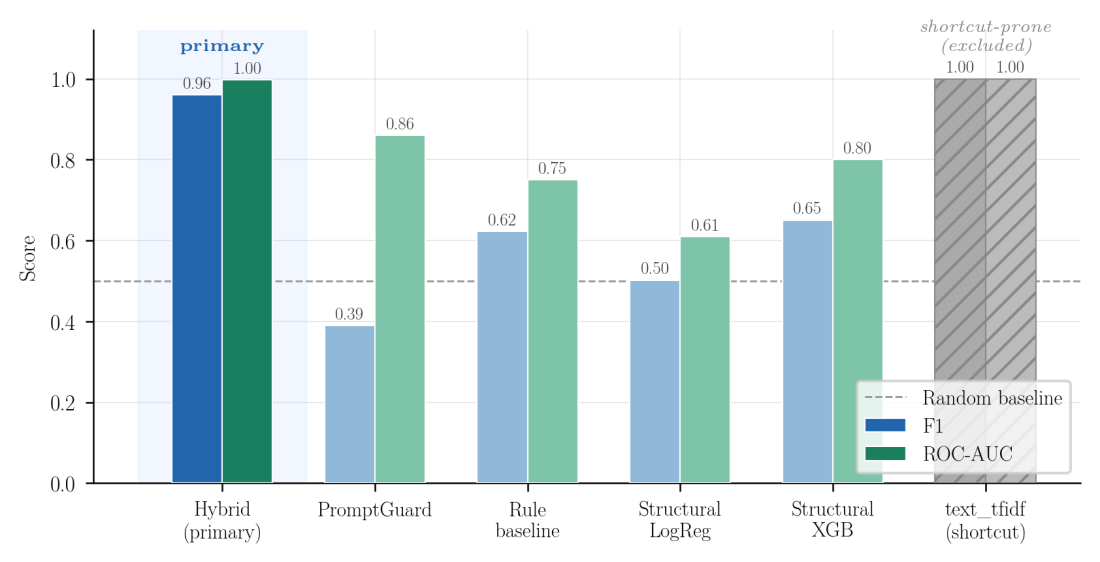}}
  \caption{Main held-out hard-provenance results. The sanitized hybrid detector is the primary model. PromptGuard, the rule baseline, and structural-only models are weaker in this setting. The text-only TF-IDF comparator reaches perfect scores but is marked shortcut-prone.}
  \label{fig-main-results}
\end{figure}

\subsection{Held-out classification performance}

On the held-out test split of 2,919 documents, the sanitized hybrid detector achieved 0.973 accuracy and 0.960 F1. Its ROC-AUC was 0.998 and its PR-AUC was 0.997. The confusion matrix contained 1,868 true negatives, 78 false positives, 2 false negatives, and 971 true positives.
The baselines were weaker. PromptGuard reached 0.390 F1 with recall 0.252 when given extracted text only. The rule baseline reached 0.623 F1. Structural-only logistic regression reached 0.502 F1, while structural-only XGBoost reached 0.651 F1. Text-only TF-IDF reached perfect held-out metrics, but the shortcut audit prevents treating that result as clean evidence.

\subsection{Label-shuffle sanity check}

To test whether performance reflected real structure rather than accidental separability or leakage, we repeated training with randomized labels. Under this control, the hybrid detector fell to F1 0.500 and ROC-AUC 0.507. Structural-only logistic regression fell to F1 0.499 and ROC-AUC 0.481. Structural-only XGBoost fell to F1 0.497 and ROC-AUC 0.512.
This result supports the interpretation that the original labels contain real signal. The detector is not simply exploiting random feature-table artifacts. The grouped split also reduces the risk that the model is memorizing related variants of the same base document across train and test.

\subsection{Shortcut-removal tests}

The shortcut audit changed the interpretation of the results. The primary hybrid model passed the wrapper-token audit with zero wrapper-token hits. Text-only TF-IDF reached perfect held-out metrics but failed shortcut framing because wrapper and synthetic tokens appeared among high-weight evidence. This is why TF-IDF is treated as a comparator rather than as the main detector.

\subsection{Paired benign-confounder evaluation}

\begin{figure}[t]
  \centering
  \makebox[\linewidth][c]{\includegraphics[width=1.25\linewidth]{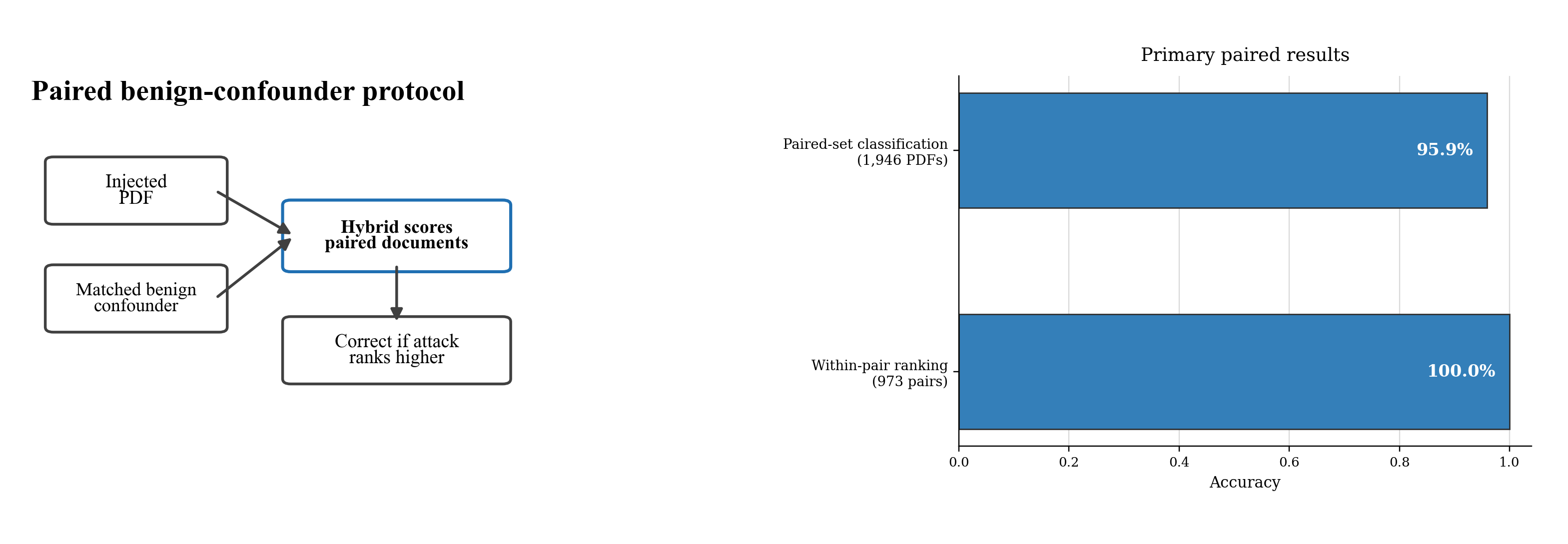}}
  \caption{Paired benign-confounder evaluation. On the balanced 1,946-PDF paired subset, the hybrid detector achieves 95.9\% thresholded classification accuracy; within-pair score ordering is correct for all 973 injected-confounder pairs (100\%).}
  \label{fig-paired}
\end{figure}

Random negatives can make the task too easy. On the balanced paired subset of 973 injected PDFs and 973 matched benign confounders, the hybrid detector achieves 95.9\% thresholded classification accuracy. Separately, its score for the injected PDF exceeds its matched confounder in all 973 pairs, yielding 100\% within-pair ranking accuracy. These results distinguish thresholded classification from relative score ordering and reduce the risk that the detector is learning only broad generation artifacts.

\subsection{Held-out attack-family stress tests}

\begin{figure}[t]
  \centering
  \makebox[\linewidth][c]{\includegraphics[width=1.25\linewidth]{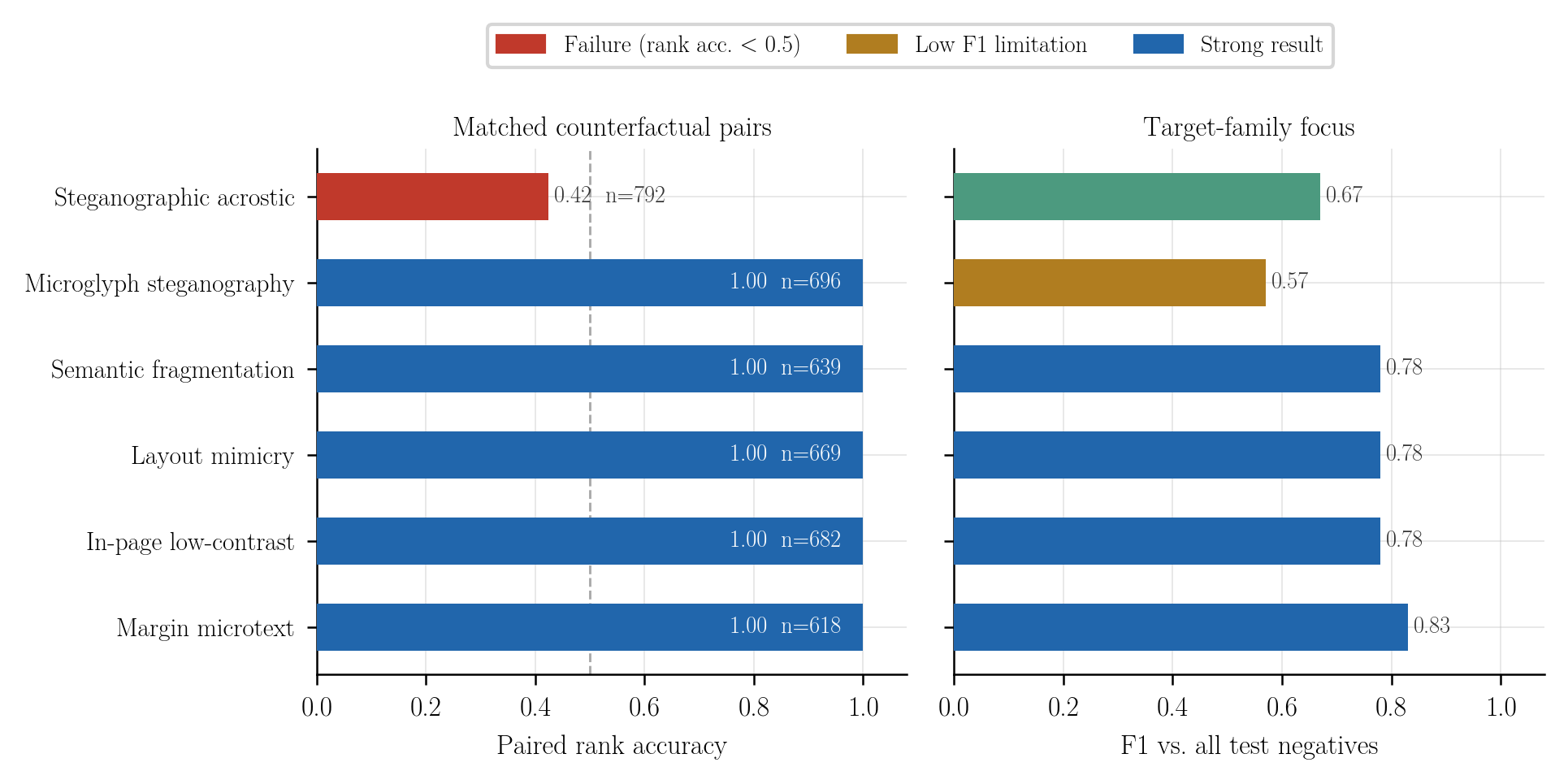}}
  \caption{Held-out attack-family stress tests. Several families remain easy for the hybrid detector, but steganographic and microglyph variants expose limits. These tests are stress evidence rather than the primary success claim.}
  \label{fig-family-stress}
\end{figure}

Held-out family tests show where the result should not be overclaimed. The hybrid detector performed well on several matched families, but steganographic acrostics had paired-rank accuracy 0.424. Microglyph steganography had perfect rank accuracy but low target-family F1 against all test negatives. These cases show that broad cross-family generalization remains open.

\subsection{Interpretation and limits}

These experiments show that document-aware detection is useful for this controlled detection task. Under paired controls, the hybrid detector can identify many injected documents before model-level guardrails or downstream retrieval.
These experiments do not establish robustness to arbitrary real-world PDFs or unseen document sources. They also do not establish robustness to unseen injection generators. OCR-only pipelines and adaptive attackers require separate evaluation. The right interpretation is that controlled document-layer injections leave detectable evidence, but the evidence is not purely structural and not universal.

\section{Conclusion}

This work studied hidden prompt injection as a document-ingestion problem rather than only a model-alignment problem. We built CrackedPDFs, a controlled PDF benchmark with paired benign confounders and hard-provenance splits. We then evaluated whether document-aware detectors could identify injected files before their contents reached an LLM pipeline.

The main result is not that every simple detector works. PromptGuard had low recall when restricted to extracted text. Structural-only models were weak under paired controls. Text-only TF-IDF reached perfect held-out scores but failed shortcut framing. The sanitized hybrid detector was the strongest clean result, with 0.960 F1 on the held-out test set, 95.9\% classification accuracy on the balanced paired subset, and 100\% within-pair ranking accuracy across 973 injected-confounder pairs.

The conclusion is limited. Controlled document-layer prompt injections can leave detectable evidence, but broad generalization is not solved. OCR-only pipelines, adaptive attackers, and arbitrary real-world PDFs require separate evaluation. The practical lesson is that document structure belongs in the security boundary for LLM systems. Once a PDF is flattened into plain text chunks, important evidence about visibility and placement can disappear. A practical defense will likely need document-aware screening at ingestion combined with downstream semantic and model-level safeguards.

\section*{Code and Data Availability}

The frozen paper release, evaluation code, exact result tables, and reproduction commands are available at \url{https://github.com/volkthienpreecha/crackedpdfs/releases/tag/v1.0.0-paper}. The complete dataset is available at \url{https://huggingface.co/datasets/volkthienpreecha/crackedpdfs} and archived at Zenodo under DOI \href{https://doi.org/10.5281/zenodo.21735803}{10.5281/zenodo.21735803}. The published results can be reproduced without regenerating the 29,322 PDFs using \texttt{make reproduce-results}. Benign source documents were generated using PDFAutoGen.

\section*{AI Usage Disclosure}

AI tools were used for sentence-level editing, \LaTeX{} formatting assistance, and coding assistance during implementation and debugging. The authors take full responsibility for the content of this manuscript.

\end{document}